\documentclass[10pt,twocolumn,letterpaper]{article}

\usepackage{iccv,balance}
\usepackage{times}
\usepackage{epsfig}
\usepackage{graphicx}
\usepackage{amsmath}
\usepackage{amssymb}
\usepackage{nicefrac}       
\usepackage{microtype}      
\usepackage{xcolor}         
\usepackage{boldline}
\usepackage{algorithm}
\usepackage{listings}
\usepackage{bm}
\usepackage{makecell}

\usepackage[colorlinks=true,linkcolor=blue,urlcolor=blue,citecolor=purple,breaklinks=true,bookmarks=false]{hyperref}

\iccvfinalcopy 

\def\iccvPaperID{****} 

\ificcvfinal\fi

\begin{document}

\title{S$^2$-MLP: Spatial-Shift MLP Architecture  for Vision}

\author{Tan Yu, Xu Li, Yunfeng Cai, Mingming Sun,  Ping Li\\
Cognitive Computing Lab\\
Baidu Research\\
10900 NE 8th St. Bellevue, Washington 98004, USA\\
No.10 Xibeiwang East Road, Beijing 100193, China\\
{\tt\small \{tanyu01,lixu13,caiyunfeng,sunmingming01,liping11\}@baidu.com}
}

\maketitle

\ificcvfinal\thispagestyle{empty}\fi

\begin{abstract}

\noindent Recently, visual Transformer (ViT) and its following works abandon the convolution and  exploit the self-attention  operation, attaining a comparable or even higher accuracy than CNNs. More recently, MLP-Mixer abandons both the convolution and the self-attention operation, proposing an architecture containing only MLP layers. To achieve cross-patch communications, it devises  an additional  token-mixing MLP besides the channel-mixing MLP. It achieves promising results when training on an extremely large-scale dataset. But it cannot achieve as outstanding  performance as its CNN and ViT counterparts when training on medium-scale datasets such as ImageNet1K and ImageNet21K. The performance drop of MLP-Mixer motivates us to rethink the token-mixing MLP. We discover that the token-mixing MLP  is a variant of the depthwise convolution  with a global reception field and  spatial-specific configuration. But the  global reception field and  the spatial-specific property  make token-mixing MLP prone to over-fitting. In this paper, we propose a novel pure MLP architecture, spatial-shift MLP (S$^2$-MLP). Different from MLP-Mixer, our S$^2$-MLP only contains channel-mixing MLP. We utilize a spatial-shift operation for  communications between patches. It has a local reception field and is spatial-agnostic. It is parameter-free and efficient for computation. The proposed S$^2$-MLP attains higher recognition accuracy than MLP-Mixer when training on ImageNet-1K dataset. Meanwhile,  S$^2$-MLP accomplishes as excellent performance as   ViT  on ImageNet-1K dataset with considerably simpler architecture and fewer FLOPs and parameters.
\end{abstract}

\section{Introduction}

In the past years, convolutional neural networks (CNN)~\cite{krizhevsky2012imagenet,he2016deep} have achieved great success in computer vision.  Recently, inspired by the triumph achieved by Transformer~\cite{vaswani2017attention} in natural language processing, visual Transformer (ViT)~\cite{dosovitskiy2020image} is proposed.  It replaces the convolution operation in CNN with the self-attention operation used in Transformer to model the visual relations between local patches in different spatial locations of the image.  
 ViT and the followup works~\cite{touvron2020training,yuan2021tokens,wang2021pyramid,liu2021swin,han2021transformer,wu2021cvt,touvron2021going} have achieved comparable or even better performance than CNN models. 
Compared with CNN  demanding a meticulous design for the convolution kernel, ViT simply stacks a series of standard Transformer blocks with identical settings, taking less hand-crafted manipulation and  reducing the inductive biases. 

More recently, MLP-Mixer~\cite{tolstikhin2021mlp} proposes a simpler alternative based entirely on multi-layer perceptrons (MLP) to further reduce the inductive biases. 
The basic block in MLP-Mixer consists of two components: the channel-mixing MLP and the  token-mixing MLP.   The channel-mixing MLP  projects the feature map along the channel dimension for the communications between different channels. In parallel,  the token-mixing MLP projects the feature map along the spatial dimension and  achieves  the communications between spatial locations.  When trained on the huge-scale dataset such as JFT-300M~\cite{sun2017revisiting}, MLP-Mixer attains promising recognition accuracy.  But there is still an accuracy gap between MLP-Mixer and ViT on medium-scale datasets,  ImageNet-1K and ImageNet-21K~\cite{deng2009imagenet}.  Specifically, Mixer-Base-16~\cite{tolstikhin2021mlp} achieves only a $76.44\%$ top-1 accuracy on ImageNet-1K, whereas ViT-Base-16~\cite{dosovitskiy2020image} achieves a $79.67\%$ top-1 accuracy.

\begin{figure}
\hspace{-0.25in}
\includegraphics[scale=0.52]{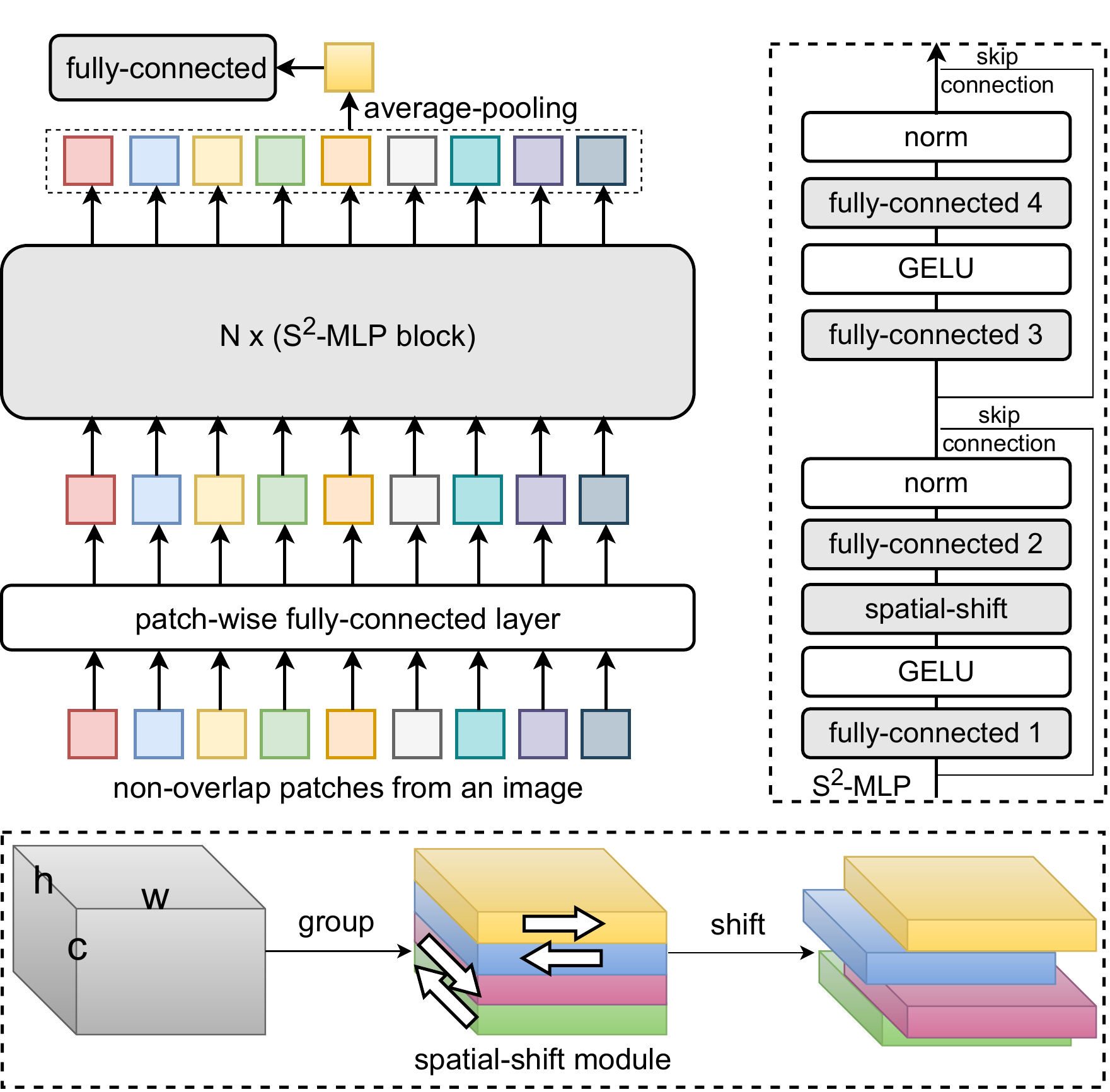}

\vspace{0.1in}

\caption{The architecture of the proposed spatial-shift multi-layer perceptions (S$^2$-MLP) model. Non-overlap patches cropped from an image are the input of the model. They go through a stack of S$^2$-MLP blocks which are further aggregated into a single feature vector through global average pooling. After that, the feature vector is  fed into a fully-connected layer for predicting the label.  An S$^2$-MLP block  contains four fully-connected layers, two GELU layers~\cite{hendrycks2016gaussian}, two layer normalization~\cite{ba2016layer}, two skip connections~\cite{he2016deep}, and a spatial-shift module. The proposed spatial-shift module groups $c$ channels into several groups. Then it shifts different groups of channels in different directions.}
\label{arch}
\vspace{-0.1in}
\end{figure}

The unsatisfactory performance of MLP-Mixer on ImageNet-1K and ImageNet-21K motivates us to rethink the mixing-token MLP in MLP-Mixer. Given $N$ patch features in the matrix form, $\mathbf{X} = [\mathbf{x}_1,\cdots,\mathbf{x}_N]$, the token-mixing MLP conducts  $\mathbf{X}\mathbf{W}$ where $\mathbf{W} \in \mathbb{R}^{N\times M}$ is the learnable weight matrix.  It is straightforward to observe that each column of $\mathbf{X}\mathbf{W}$, the output of the token-mixing MLP, is a weighted summation of patch features (columns in the input $\mathbf{X}$). The weights in summation are similar to the attention in Transformer.  But the attention in Transformer is data-dependent, whereas the weights for summation in token-mixing MLP are agnostic to the input. To some extent, the weighted summation is more similar to depthwise  convolution~\cite{chollet2017xception,howard2017mobilenets,kaiser2018depthwise}. But the depthwise convolution  only has a local reception field. In contrast, token-mixing MLP has a global reception field. Besides, the depthwise  convolution kernel is shared among different spatial locations, whereas the weights for summation in token-mixing MLP are different for different spatial locations. Without the limitation of  the local reception field and the spatial-agnostic constraint, the token-mixing MLP is more flexible and has a stronger fitting capability than the depthwise convolution. But the freedom from breaking a chain is accompanied by the risk of over-fitting. To avoid suffering from the over-fitting  in token-mixing MLP, a huge number of training samples should be provided. It explains the fact that after pre-training on the ultra large-scale dataset, JFT-300M,  the recognition accuracy gap between MLP-Mixer and ViT is shortened.


To alleviate the over-fitting issue of MLP-Mixer when only medium-scale training data is available, we propose a  spatial-shift MLP (S$^2$-MLP) architecture, a conceptually simple architecture containing only channel-mixing MLPs. To conduct communication between  spatial locations, we adopt a spatial-shift operation, which is parameter-free and  efficient for computation. The spatial-shift operation is spatial-agnostic and  maintains a local reception field. Figure~\ref{arch} illustrates the architecture of the proposed S$^2$-MLP. It crops an image into $w\times h$ non-overlap patches. For each patch, it obtains the patch embedding vector through a fully-connected layer. The patch embedding vectors further  go through $N$ S$^2$-MLP blocks.  Each S$^2$-MLP block contains four fully-connected layers. The fully-connected layer  in  each S$^2$-MLP block serves as the same function as the channel-mixing MLP used in MLP-Mixer. But our S$^2$-MLP does not need token-mixing MLP. Instead, the communications between different spatial locations  are achieved through the proposed spatial-shift module.  It is parameter-free and simply shifts channels from a patch to its adjoining patches. Despite that the spatial-shift module  only supports communications between adjacent patches, stacking a series of  S$^2$-MLP blocks makes the long-range communications feasible. 

The proposed S$^2$-MLP is deceptively simple in architecture. It attains considerably higher recognition accuracy than MLP-Mixer on ImageNet1K dataset with a comparable scale of parameters and FLOPs. Meanwhile, it achieves a comparable recognition accuracy with respect to ViT on ImageNet1K dataset with a considerably simpler structure, fewer parameters and FLOPs.

\section{Related Work}

\noindent \textbf{Transformer-based vision models.}\; Visual Transformer (ViT)~\cite{dosovitskiy2020image} is the first work to build a purely Transformer-based vision backbone. Through training on an extremely large-scale dataset, JFT-300M~\cite{sun2017revisiting}, it has achieved promising results compared with \emph{de facto} vision backbone, convolutional neural network.  DeiT~\cite{touvron2020training} adopts the advanced training and augmentation strategy and achieves excellent performance when trained on ImageNet-1K only. Recently, several works further improve the performance of visual Transformer from multiple perspectives. For instance, PVT~\cite{wang2021pyramid} uses a progressive shrinking pyramid to reduce computations of large feature maps. T2T~\cite{yuan2021tokens} progressively tokenizes the image to model the local structure information of the image. TNT~\cite{han2021transformer} constructs another Transformer within the outer-level Transformer to model the local patch. CPVT~\cite{chu2021conditional} proposes a  conditional positional encoding to effectively encode the spatial locations of patches.  Visual Longformer~\cite{zhang2021multi} adopts the global tokens to  boost efficiency. PiT~\cite{heo2021pit} investigates the spatial dimension conversion and integrates pooling layers between self-attention blocks.  Swin-Transformer~\cite{liu2021swin} adopts a  hierarchical architecture of high flexibility to model the image at various scales. Twins~\cite{chu2021twins}  utilizes a hierarchical structure consists of a locally-grouped self-attention and a global sub-sampled attention. CaiT~\cite{touvron2021going} builds and optimizes deeper transformer networks for image classification. 

\vspace{0.1in}
\noindent \textbf{MLP-based vision models.}\;
MLP-Mixer~\cite{tolstikhin2021mlp} proposes a conceptually  and technically simple architecture solely based on MLP layers. To model the communications between spatial locations, it proposes a token-mixing MLP. Despite that MLP-Mixer has achieved promising results when trained on a huge-scale dataset JFT-300M, it is not as good as its visual Transformer counterparts when trained on a medium-scale dataset including ImageNet-1K and ImageNet-21K.
FF~\cite{melas2021you} adopts a similar architecture but inherits the global [CLS] token and positional embedding from ViT. Res-MLP~\cite{touvron2021resmlp} also designs a pure MLP architecture. It proposes an affine transform layer which facilities stacking a huge number of MLP blocks. Using a deeper architecture than MLP-Mixer, Res-MLP  achieves a higher accuracy than MLP-Mixer and a comparable recognition accuracy as ViT.
gMLP~\cite{liu2021pay} designs a gating operation to enhance the communications between spatial locations and achieves a comparable recognition accuracy compared with  DeiT. EA~\cite{guo2021beyond} replaces the self-attention module with an external attention through external  memories learned from the training data. It is implemented by a cascade of two linear layers.  Container~\cite{gao2021container} proposes  a generalized context aggregation building block that combines static affinity matrices as token-mixer and dynamic affinity matrices as visual Transformers.

\section{Method}

\noindent In this section, we describe spatial-shift MLP (S$^2$-MLP).   

\subsection{Preliminary}

\vspace{0.05in}

\noindent \textbf{Layer Normalization (LN)}~\cite{ba2016layer} is a widely used for models using Transformer and BERT architecture.  Given a $c$-dimensional vector $\mathbf{x} = [x_1,\cdots, x_c]$, layer normalization computes  the mean $\mu =  \frac{1}{c}\sum_{i=1}^c x_i$ and the standard deviation $\sigma = \sqrt{\frac{1}{c}\sum_{i=1}^c (x_i - \mu)^2}$.  It normalizes each entry in $\mathbf{x}$ by $\bar{x}_i = \gamma \frac{x_i - \mu}{\sigma} + \beta$, where $\beta$ and $\gamma$ are learnable parameters.

\vspace{0.15in}
\noindent \textbf{Gaussian Error Linear Units (GELU)}~\cite{hendrycks2016gaussian} is a widely used activation function in Transformer and BERT models.   It is  defined as 
$\mathrm{GELU}(x) = x \Phi (x)$, 
where $\Phi (x)$ is the standard Gaussian cumulative distribution function defined as $\Phi (x) = \frac{1}{2}[1+\mathrm{erf}(x/\sqrt{2})].$

\vspace{0.15in}
\noindent \textbf{MLP-Mixer}~\cite{tolstikhin2021mlp} stacks $N$ basic blocks of identical size and structure. Each basic block consists of two types of MLP layers:   channel-mixing MLP and token-mixing MLP.   Let us denote a patch feature by $\mathbf{p}_i \in \mathbb{R}^c$ and an image with $n$ patch features by $\mathbf{P} = [\mathbf{p}_1,\cdots,\mathbf{p}_n] \in \mathbb{R}^{c\times n}$.  Channel-mixing MLP projects $\mathbf{P}$ along the channel dimension:
\begin{equation}
\hat{\mathbf{P}} = \mathbf{P} + \mathbf{W}_2 \mathrm{GELU}(\mathbf{W}_1\mathrm{LN}(\mathbf{P})),
\end{equation}
 where $\mathbf{W}_1 \in \mathbb{R}^{\hat{c} \times c}$ and  $\mathbf{W}_2 \in \mathbb{R}^{{c} \times \hat{c}}$. 
  In parallel, token-mixing  MLP projects the channel-mixed  patch features $\hat{\mathbf{P}}$ along the spatial dimension:
 \begin{equation}
\bar{\mathbf{P}} = \hat{\mathbf{P}} + \mathrm{GELU}(\mathrm{LN}(\hat{\mathbf{P}})\mathbf{W}_3) \mathbf{W}_4 ,
\end{equation}
 where $\mathbf{W}_3 \in \mathbb{R}^{N\times \bar{N}}$ and $\mathbf{W}_4 \in \mathbb{R}^{\bar{N}\times {N}}$.

\vspace{0.1in}

\subsection{Spatial-Shift MLP Architecture}

 As shown in Figure~\ref{arch}, our spatial-shift MLP backbone consists of a path-wise fully-connected layer, $N$ S$^2$-MLP blocks, and a fully-connected layer for classification. Since we have introduced   the fully-connected layer for classification is well-known, we only introduce patch-wise fully-connected layer and  the proposed spatial-shift block. The proposed spatial-shift operation is closely related to  Shift~\cite{wu2018shift}, 4-connected Shift~\cite{brown20194} and TSM~\cite{lin2019tsm}. Our spatial-shift operation can be regarded as a special version of 4-Connected Shift without origin element information. Different from the 4-connected shift residual block~\cite{brown20194} in a fc-shift-fc structure, our S$^2$-MLP block, as visualized in Figure~\ref{arch},  takes another two fully-connected layers only for mixing channels after a fc-shift-fc structure. Besides, 4-connected shift residual network exploits  convolution in the early layer, whereas ours adopts a pure-MLP structure.

\vspace{0.15in}
\noindent \textbf{Patch-wise fully-connected layer.}\; 	We denote an image by ${I} \in \mathbb{R}^{W\times H \times 3}$. It is uniformly split into  $w\times h$ patches, $\mathcal{P} = \{\bm{\mathcal{P}}_i\}_{i=1}^{wh}$, where $\bm{\mathcal{P}}_i \in \mathbb{R}^{p\times p\times 3}$, $w = \frac{W}{p}$, and $h = \frac{H}{p}$.  For each patch $\bm{\mathcal{P}}_i$, we unfold it into a vector $\mathbf{p}_i \in \mathbb{R}^{3p^2}$ and  project it into an embedding vector $\mathbf{e}_i$ through a fully-connected layer followed by a layer normalization:
\begin{equation*}
\mathbf{e}_i = \mathrm{LN}(\mathbf{W}_0 \mathbf{p}_i  + \mathbf{b}_0), 
\end{equation*}
where $\mathbf{W}_0 \in \mathbb{R}^{c\times 3p^2}$ and $\mathbf{b}_0 \in \mathbb{R}^{c}$ are parameters of the fully-connected layer and $\mathrm{LN}(\cdot)$ denotes the layer normalization which we will have introduced above.

\vspace{0.15in}\noindent \textbf{S$^2$-MLP block.}\;  Our architecture stacks $N$ S$^2$-MLP  of the same size and structure. Each spatial-shift block contains four fully-connected layers, two layer-normalization layers, two GELU layers, two skip-connections, and the proposed spatial-shift module. It is worth noting that all fully-connected layers used in our S$^2$-MLP only serve to mix the channels.  We do not use the token-mixing MLP in MLP-Mixer. Since the fully-connected layer is well known, and we have already introduced layer normalization and GELU above, we only focus on the proposed  spatial-shift module here.
We denote the feature map in the input of our  spatial-shift module by $\bm{\mathcal{T}} \in \mathcal{R}^{w\times h \times c}$, where $w$ denotes the width, $h$ represents  the height, and $c$ is the number of channels.   The spatial-shift operation can be decomposed into two steps: 1)  split the channels into several groups, and 2) shift each group of channels in different directions.

\vspace{0.1in} \noindent  \textbf{Group.}\; We uniformly split $\bm{\mathcal{T}}$ along the channel dimension and obtain $g$ thinner tensors $\{\bm{\mathcal{T}}_{\tau} \}_{\tau=1}^g$ where $\bm{\mathcal{T}}_{\tau} \in \mathcal{R}^{w\times h \times c/g}$.
It is worth noting that the number of groups, $g$, is dependent on the design of the shifting directions in the second step. For instance, by default, we only  shift along four directions, and thus $g$ is set as $4$ in this configuration.

\vspace{0.1in} \noindent  \textbf{Spatial-shift operation.}\;    We shift different groups in different directions. For the first group of channels, $\bm{\mathcal{T}}_1$, we shift it along the wide dimension by $+1$. In parallel, we shift the second group of channels, $\bm{\mathcal{T}}_1$, along the wide dimension by $-1$.  Similarly, for $\bm{\mathcal{T}}_3$, we shift it along the height dimension by $+1$, and  we shift  $\bm{\mathcal{T}}_4$  along the height dimension by $-1$. We clarify the formulation of the spatial-shift operation in Eq.~\eqref{eq:shift} and demonstrate the pseudocode in Algorithm~\ref{alg:code}.
\begin{equation}
\label{eq:shift}
\begin{split}
&\bm{\mathcal{T}}_1[1:w,:,:] \gets  \bm{\mathcal{T}}_1[0:w-1,:,:],\\
&\bm{\mathcal{T}}_2[0:w-1,:,:] \gets  \bm{\mathcal{T}}_2[1:w,:,:],\\
&\bm{\mathcal{T}}_3[:,1:h,:] \gets  \bm{\mathcal{T}}_3[:,0:h-1,:],\\
&\bm{\mathcal{T}}_4[:,0:h-1,:] \gets  \bm{\mathcal{T}}_4[:,1:h,:].\\
\end{split}
\end{equation}
After spatially shifting, each patch absorbs  the visual content from its adjoining patches. The spatial-shift operation is parameter-free and makes the communication between different spatial locations feasible.  The above mentioned spatial-shift manner is one of simplest and most straightforward methods for shifting. We also evaluate other spatial-shift manners. 
Surprisingly, the above simple manner has achieved excellent performance compared with others. Using the spatial-shift operation, we no longer need token-mixer as MLP-Mixer. We only need channel-mixer to project the patch-wise feature along the channel dimension. Note that the spatial-shift operation in a single block is only able to gain the visual content from adjacent patches and cannot have access to visual content of all patches in the image. But we stack $N$ S$^2$-MLP blocks, the global visual content will be gradually diffused to every patch.

\subsection{Relations with depthwise convolution}
\label{limdir}

\begin{algorithm}[t]
\caption{Pseudocode of our spatial-shift operation.}
\label{alg:code}
\lstset{
  backgroundcolor=\color{white},
  basicstyle=\fontsize{9.5pt}{9.5pt}\ttfamily\selectfont,
  columns=fullflexible,
  breaklines=true,
  captionpos=b,
  commentstyle=\fontsize{9.5pt}{9.5pt}\color{codeblue},
  keywordstyle=\fontsize{9.5pt}{9.5pt},
}
\begin{lstlisting}[language=Python]
def spatial_shift(x):
  w,h,c = x.size()
  x[1:,:,:c/4] = x[:w-1,:,:c/4]
  x[:w-1,:,c/4:c/2] = x[1:,:,c/4:c/2]
  x[:,1:,c/2:c*3/4] = x[:,:h-1,c/2:c*3/4]
  x[:,:h-1,3*c/4:] = x[:,1:,3*c/4:]
  return x 
\end{lstlisting}
\end{algorithm}

\noindent \textbf{Depthwise convolution.}\; 
Given a feature map defined as a tensor $\mathcal{T} \in \mathbb{R}^{w\times h \times c}$, depthwise convolution~\cite{chollet2017xception,howard2017mobilenets,kaiser2018depthwise} utilize a two dimensional convolution kernel $\mathbf{K}_i$ separably on  each two-dimensional slice of the tensor $\mathcal{T}[:,:,i] \in \mathbb{R}^{w\times h}$ where $i \in [1,c]$.
Depthwise convolution takes cheap computational cost and thus is widely used in efficient neural network for fast inference. 


\noindent \textbf{Relations.}\; 
In fact, the spatial-shift operation is equivalent to a depthwise  convolution with a fixed and group-specific kernel weights. Let denote a set of depthwise  convolution kernels as $\mathcal{K} = \{\mathbf{K}_1, \cdots, \mathbf{K}_c \}$. If we set 
\begin{align*}
\label{kernel}
 \mathbf{K}_i &=  
\begin{bmatrix}
0  & 0 & 0\\
1 &  0 & 0 \\
0  & 0 & 0 
\end{bmatrix}, \quad \forall  i\in(0,\frac{c}{4}],  \\
\mathbf{K}_j &= \begin{bmatrix}
0  & 0 & 0\\
0 &  0 & 1 \\
0  & 0 & 0 
\end{bmatrix}, \quad \forall j\in (\frac{c}{4},\frac{c}{2}], \\
 \mathbf{K}_k &=  
\begin{bmatrix}
0  & 1 & 0\\
0 &  0 & 0 \\
0  & 0 & 0 
\end{bmatrix}, \quad \forall k\in (\frac{c}{2},\frac{3c}{4}],\\
\mathbf{K}_l &= \begin{bmatrix}
0  & 0 & 0\\
0 &  0 & 0 \\
0  & 1 & 0 
\end{bmatrix}, \quad \forall  l\in (\frac{3c}{4},c],
\end{align*}
 the depthwise convolution based on the group of kernels $\mathcal{K}$ is equivalent to our spatial-shift operation. 

That is, our spatial-shift operation is a variant of the depthwise  convolution with the fixed weights defined above.  Meanwhile, the spatial-shift operation shares kernel weights within each group of channels. As  mentioned in the introduction section, the token-mixing MLP in MLP-Mixer is a global-reception and spatial-specific variant of the depthwise  convolution. Different from our spatial-shift operation and vanilla depthwise  convolution, the weights for summation in token-mixing are shared cross channels for a specific spatial location. In contrast, the vanilla depthwise convolution learns 
different convolution kernels for different channels, and our spatial-shift operation shares the weights within the group and adopts different weights for different~groups.

 \begin{table}[b!]
\vspace{-0.05in}
 
\centering
\begin{tabular}{ccccc}
\hlineB{3}
 & weights & \makecell{reception \\ field} & \makecell{spatial} & \makecell{channel} \\ \hline
TM  &  learned   & global     &   specific     &   agnostic    \\
S$^{2}$ &   fixed   & local  &    agnostic    &     group-specific  \\
DC &   learned   &   local  & agnostic       &  specific     \\\hlineB{3}
\end{tabular}
\vspace{0.1in}
\caption{Relations among token-mixing (TM),   spatial-shift (S$^2$) and  depthwise convolution (DC).}
\label{rela}
\end{table}

We  summarize their relations and differences in Table~\ref{rela}. Observing the connections between  the spatial-shift operation and  depthwise  convolution, we encourage the researchers to attempt depthwise  convolution with different settings to build new MLP-based architectures.

\vspace{0.1in}

\subsection{Complexity Analysis}

\vspace{0.05in}
\noindent \textbf{Patch-wise fully-connected layer (PFL)} projects each  patch cropped from the raw image,  $\bm{\mathcal{P}} \in \mathbb{R}^{p\times p \times 3}$,  into a $c$-dimensional  feature vector. The weights of PFL satisfy $\mathbf{W}_0 \in \mathbb{R}^{c \times 3p^2 }$ and $\mathbf{b}_0 \in \mathbb{R}^{c}$. Thus, the number of parameters in PFL is 
\begin{equation*}
    \mathrm{Params}_{\mathrm{PFL}} = (3p^2+1)c.
\end{equation*}
The total number of patches is $M = w \times h =  \frac{W}{p} \times \frac{{H}}{p}$ where $W$ is the width and $H$ is the height of the input image. In this case, the floating operations (FLOPs) in PFL is 
\begin{equation*}
  \mathrm{FLOPs}_{\mathrm{PFL}} =  3Mcp^2. 
\end{equation*}
It is worth noting that,  following previous works~\cite{touvron2020training,han2021transformer},  we only consider the multiplication  operation between float numbers when counting FLOPs.

\vspace{0.1in} \noindent \textbf{S$^2$-MLP blocks.}\; The proposed  S$^2$-MLP vision architecture consists of $N$ S$^2$-MLP blocks. The input and output of all blocks  are of the same size.
We denote the input of the $i$-th S$^2$-MLP block by an tensor $\bm{\mathcal{T}}_{\mathrm{in}}^{(i)}$ and the output by $\bm{\mathcal{T}}_{\mathrm{out}}^{(i)}$.  Then, these tensors satisfy
\begin{equation*}
    \bm{\mathcal{T}}_{\mathrm{in}}^{(i)}, \bm{\mathcal{T}}_{\mathrm{out}}^{(i)} \in \mathbb{R}^{w\times h \times c}, \quad \forall i \in [1,N].
\end{equation*}
All S$^2$-MLP  blocks take the same operation and are of the same configuration.
This leads to the fact that all blocks take the same computational cost and the same number of parameters. To obtain the total number of parameters and FLOPs of the proposed S$^2$-MLP architecture, we only need count that for each basic block.

Only fully-connected layers contain parameters. As shown in Figure~\ref{arch}, S$^2$-MLP contains four fully-connected layers.   We denote the weights of the first two fully-connected layer as $\{\mathbf{W}_1,\mathbf{b}_1\}$ and $\{\mathbf{W}_2,\mathbf{b}_2\}$ where 
$\mathbf{W}_1 \in \mathbb{R}^{c\times c}$ and 
$\mathbf{W}_2 \in \mathbb{R}^{c \times c}$. These two fully-connected layers keep the feature dimension unchanged.
We denote the weights of the third fully-connected layer as $\{\mathbf{W}_3, \mathbf{b}_3\}$ where $\mathbf{W}_3 \in \mathbb{R}^{\bar{c} \times c}$ and 
$\mathbf{b}_3 \in \mathbb{R}^{\bar{c}}$. $\bar{c}$ denotes the hidden size. Following ViT and MLP-Mixer, we set $\bar{c} = rc$ where $r$ is the expansion ratio which is set as $4$, by default. In this step, the feature dimension of each patch increases from $c$ to $\bar{c}$.  In contrast, the fourth fully-connected layer reduces the dimensionality  of each patch feature from $\bar{c}$ back to $c$. The dimensions for the weights are $\mathbf{W}_4 \in \mathbb{R}^{c \times \bar{c}}$
and~$\mathbf{b}_4 \in \mathbb{R}^{c}$.

Therefore, the number of parameters per S$^2$-MLP block  is the total number of entries in $\{\mathbf{W}_i,\mathbf{b}_i\}_{i=1}^4$ is
\begin{equation*}
    \mathrm{Params}_{\mathrm{S}^{2}} = c(2c+2\bar{c}) + 3c + \bar{c} = c^2(2r+2)+ c(3+r),
\end{equation*}
and the total FLOPs of fully-connected layers in each  S$^2$-MLP block becomes
\begin{equation*}
    \mathrm{FLOPs}_{\mathrm{S}^2} =  M(2c^2 + 2c\bar{c}) = Mc^2(2r+2).
\end{equation*}

\noindent \textbf{Fully-connected classification layer (FCL)} takes input the $c$-dimensional vector from average-pooling $M$ patch features in the output of the last S$^2$-MLP block. It outputs $k$-dimensional score vector where $k$ is the number of classes. Hence, the number of parameters in FCL is
\begin{equation*}
    \mathrm{Params}_{\mathrm{FCL}} = (c+1)k.
\end{equation*}
The FLOPs of FCL is
\begin{equation*}
    \mathrm{FLOPs}_{\mathrm{FCL}} = Mck.
\end{equation*}
By adding up the number of parameters in the patch-wise fully-connected layer, $N$ S$^2$-MLP blocks, and the fully-connected classification layer, we obtain the total number of parameters of the entire architecture:
\begin{equation*}
  \mathrm{Params} =     \mathrm{Params}_{\mathrm{PFL}} + N*\mathrm{Params}_{\mathrm{S}^2} +  \mathrm{Params}_{\mathrm{FCL}}. 
\end{equation*}
Therefore the total number of FLOPs is 
\begin{equation*}
  \mathrm{FLOPs} =      \mathrm{FLOPs}_{\mathrm{PFL}} + N*\mathrm{FLOPs}_{\mathrm{S}^2} +  \mathrm{FLOPs}_{\mathrm{FCL}}. 
\end{equation*}

\vspace{0.1in}

\subsection{Implementation}
\label{imp}
We set the cropped patch size ($p\times p$) as $16\times 16$. We reshape input image into the $224\times 224$ size. Thus, the number of patches $M = (224/16)^2 = 196$. We set expansion ratio $r=4$. 
We attempt two types of settings: 1) wide settings and 2) deep settings. The wide settings follow the base model of MLP-Mixer~\cite{tolstikhin2021mlp}. The wide settings  set  the number of  S$^2$-MLP blocks ($N$) as $12$  and the hidden size $c$ as $768$.  Note that  MLP-Mixer also implements the large model and the huge model. Nevertheless, our limited computing resources cannot afford the expensive cost of investigating the large and huge models on  ImageNet-1K dataset. 
The deep settings follow ResMLP-36~\cite{touvron2021resmlp}.  The deep settings  set  the number of  S$^2$-MLP blocks ($N$) as $36$  and the hidden size $c$ as $384$. 
We summarize the  hyperparameters, the number of parameters, and FLOPs  of two settings in Table~\ref{para}.

\vspace{0.1in}

\begin{table}[htp!]
\centering
\begin{tabular}{cccccccc}
\hlineB{3}
Settings  & $M$ &$N$ & $c$ & $r$ & $p$ & Para. & FLOPs  \\\hline
wide  & $196$  & $12$ & $768$  & $4$   & $16$ & $71$M & $14$B \\
deep  & $196$  & $36$ & $384$  & $4$   & $16$ & $51$M & $10.5$B \\
  \hlineB{3}
\end{tabular}
\vspace{0.1in}
\caption{The hyper-parameters, the number of parameters and FLOPs. Following MLP-Mixer~\cite{tolstikhin2021mlp}, the number of parameters excludes the weights of the fully-connected layer for classification.}
\label{para}
\end{table}

\begin{table*}[htp!]
\centering
\begin{tabular}{c|c|c|c|c|c}
\hlineB{3}
Model & Resolution & Top-1 ($\%$) & Top5 ($\%$)  & Params (M) & FLOPs (B) \\ 
\hlineB{3}
\multicolumn{6}{c}{CNN-based}  \\ \hline
ResNet50~\cite{he2016deep}  &  $224 \times 224$    &  $76.2$    &     $92.9$   &    $25.6$ & $4.1$   \\
4-connected Shift~\cite{brown20194}  &  $224 \times 224$    &  $77.8$    &     $-$   &    $40.8$ & $7.7$   \\
ResNet152~\cite{he2016deep}  &  $224 \times 224$    &  $78.3$    &     $94.1$   &    $60.2$ & $11.5$   \\
RegNetY-8GF~\cite{radosavovic2020designing}  &  $224 \times 224$    &  $79.0$    &     $-$   &    $39.2$ & $8.0$   \\ 
RegNetY-16GF~\cite{radosavovic2020designing}  &  $224 \times 224$    &  $80.4$    &     $-$   &    $83.6$ & $15.9$   \\
EfficientNet-B3~\cite{tan2019efficientnet}  &  $300 \times 300$    &  $81.6$    &     $95.7$   &    $12$ & $1.8$  
\\
EfficientNet-B5~\cite{tan2019efficientnet}  &  $456 \times 456$    &  $84.0$    &     $96.8$   &    $30$ & $9.9$  
\\
\hlineB{3}
\multicolumn{6}{c}{Transformer-based}  \\ \hline
ViT-B/16~\cite{dosovitskiy2020image}&  $384 \times 384$    &  $77.9$    &        &     $86.4$ & $55.5$    \\
ViT-B/16$^{*}$ ~\cite{dosovitskiy2020image,tolstikhin2021mlp}&  $224 \times 224$    &  $79.7$     & $-$       &    $86.4$ & $17.6$  \\
DeiT-B/16~\cite{touvron2020training} &   $224 \times 224$   &  $81.8$    & $-$       &  $86.4$ & $17.6$     \\
PiT-B/16~\cite{heo2021pit} &   $224 \times 224$   &  $82.0$    & $-$       &  $73.8$ & $12.5$     \\
PVT-Large~\cite{wang2021pyramid} &    $224 \times 224$   &  $82.3$    & $-$       &  $61.4$ & $9.8$     \\ 
CPVT-B~\cite{chu2021conditional} &    $224 \times 224$   &  $82.3$    & $-$       &  $88$ & $17.6$     \\
TNT-B~\cite{han2021transformer} &    $224 \times 224$   &  $82.8$    & $96.3$       &  $65.6$ & $14.1$     \\
T2T-ViT$_\mathrm{t}$-24~\cite{yuan2021tokens} &    $224 \times 224$   &  $82.6$    & $-$       &  $65.1$ & $15.0$     \\
CaiT-S32~\cite{touvron2021going} &    $224 \times 224$   &  $83.3$    & $-$       &  $68$ & $13.9$     \\
Swin-B~\cite{liu2021swin} &    $224 \times 224$   &  $83.3$    & $-$       &  $88$ & $15.4$    
\\
Nest-B~\cite{zhang2021aggregating} &    $224 \times 224$   &  $83.8$    & $-$       &  $68$ & $17.9$    \\
Container~\cite{gao2021container} &    $224 \times 224$   &  $82.7$    & $-$       &  $22.1$ & $8.1$    
\\ \hlineB{3}
\multicolumn{6}{c}{MLP-based ($c=768$, $N=12$) }  \\ \hline
   Mixer-B/16~\cite{tolstikhin2021mlp} &    $224 \times 224$   &  $76.4$    & $-$       &  $59$ & $11.6$     \\
    FF~\cite{melas2021you} &    $224 \times 224$   &  $74.9$    & $-$       &  $59$ & $11.6$     \\
     S$^2$-MLP-wide~(ours) &    $224 \times 224$   &  $80.0$    & $94.8$       &  $71$ & $14.0$     \\ \hline 
     \multicolumn{6}{c}{MLP-based ($c=384$, $N=36$) }  \\ \hlineB{3}
     ResMLP-36~\cite{touvron2021resmlp} &    $224 \times 224$   &  $79.7$    & $-$       &  $45$ & $8.9$     \\

            S$^2$-MLP-deep~(ours) &    $224 \times 224$   &  $80.7$    & $95.4$       &  $51$ & $10.5$     \\\hlineB{3}
\end{tabular}
\vspace{0.1in}
\caption{Results of our S$^2$-MLP architecture and other models on ImageNet-1K benchmark without extra data.  ViT-B/16$^{*}$ denotes the result of ViT-B/16 model reported in MLP-Mixer~\cite{tolstikhin2021mlp} with extra regularization.}
\label{main}
\end{table*}

\section{Experiments}
\label{exp}

\noindent \textbf{Datasets.}\; We evaluate the performance of the proposed S$^2$-MLP on the widely-used public benchmark, ImageNet-1K~\cite{deng2009imagenet}. It consists of $1.2$ million training images from one thousand categories and $50$ thousand validation images with $50$ images in each category. Due to the limited computing resources, the ablation study is only  conducted on a subset of ImageNet-1K, which we term as ImageNet100.  It only contains  images of randomly selected $100$ categories. It consists of $0.1$ million training images and $5$ thousand images for validation.

\vspace{0.05in}
\noindent\textbf{Training details.}\;  We adopt the training strategy provided by DeiT~\cite{touvron2020training}. To be specific, we train our model using AdamW~\cite{loshchilov2018decoupled} with weight decay $0.05$ and a batch size of $1024$.  We use a linear warmup and cosine decay.  The initial learning rate is $1$e-$3$ and gradually drops to $1$e-$5$ in $300$ epochs. We also use label smoothing~\cite{szegedy2016rethinking}, DropPath~\cite{larsson2016fractalnet}, and repeated augmentation~\cite{hoffer2020augment}. All training is conducted on a Linux server equipped with four NVIDIA Tesla V100 GPU cards. The whole training process of S$^2$-MLP  on ImageNet-1K dataset takes around 4 days. The code will be available based on the PaddlePaddle deep learning platform.

\subsection{Main results}

The main results are summarized in Table~\ref{main}.
As shown in the table,  compared with ViT~\cite{dosovitskiy2020image}, Mixer-B/16 is not competitive in terms of accuracy.  In contrast, the proposed S$^2$-MLP has obtained a comparable accuracy with respect to ViT. Meanwhile, Mixer-B/16 and our S$^2$-MLP  take considerably fewer parameters and FLOPs, making them more attractive compared with ViT when efficiency is important. We note that, by introducing some hard-crafted design,  following Transformer-based works such as PVT-Large~\cite{wang2021pyramid},  TNT-B~\cite{han2021transformer}, T2T-ViT$_\mathrm{t}$-24~\cite{yuan2021tokens}, CaiT~\cite{touvron2021going}, Swin-B~\cite{liu2021swin}, and Nest-B~\cite{zhang2021aggregating} have considerably improved  ViT.  MLP-based models including  the proposed 
S$^2$-MLP cannot achieve as high recognition accuracy as the state-of-the-art Transformer-based vision models such as CaiT, Swin-B and Nest-B. The state-of-the-art  Transformer-base vision model,  Nest-B, cannot achieve better trade-off between the recognition accuracy and efficiency compared with the state-of-the-art CNN model, EfficientNet-B5~\cite{tan2019efficientnet}.

After that, we compare our S$^2$-MLP architecture with its MLP counterparts which are recently proposed including MLP-Mixer, FF~\cite{melas2021you}, ResMLP-36~\cite{touvron2021resmlp}.
Among them, MLP-Mixer, FF and ResMLP-36 adopt a similar structure.
A difference between ResMLP-36 and MLP-Mixer is that ResMLP-36 develops an affine transformation layer to replace the layer normalization for a more stable training. Moreover, ResMLP-36  stacks more MLP layers than MLP-Mixer but uses a smaller hidden size.
Specifically, ResMLP-36 adopts $36$ MLP layers with a $384$ hidden size. In contrast,  MLP-Mixer uses $12$ MLP layers with a $768$ hidden size. Through a trade-off between the number of MLP layers and hidden size,   ResMLP-36 leads to a  higher accuracy than MLP-Mixer but takes less parameters and FLOPs.

\vspace{0.1in}

Our wide model,  S$^2$-MLP-wide adopts the wide settings in Table~\ref{para}.  Specifically, same as   MLP-Mixer and FF,   S$^2$-MLP-wide adopts 12 blocks with hidden size $768$. As shown in Table~\ref{main}, compared with MLP-Mixer and FF, the proposed S$^2$-MLP-wide achieves a considerably higher recognition accuracy. Specifically, MLP-Mixer only achieves a $76.4\%$ top-1 accuracy and FF only achieves a $74.9\%$ accuracy. In contrast, the top-1 accuracy of the proposed S$^2$-MLP-wide is $80.0\%$. In parallel, our deep model,  S$^2$-MLP-deep adopts the deep settings in Table~\ref{para}.  
 Specifically, same as   ResMLP,   S$^2$-MLP-deep adopts 36 blocks with hidden size $384$. We also use the affine transformation proposed in ResMLP to replace layer normalization for a fair comparison.
As shown in Table~\ref{main}, compared with ResMLP-36, our  S$^2$-MLP-deep achieves  higher recognition accuracy. 
Another drawback of  MLP-Mixer and ResMLP is that, the size of the weight matrix in  token-mixer  MLP, $\mathbf{W} \in \mathbb{R}^{N\times N}$  ($N = wh$),  is dependent on the feature map size.  \textbf{That is, the structure of MLP-Mixer as well as ResMLP varies as the input scale changes.} Thus,  MLP-Mixer and ResMLP trained on the feature map of $14\times 14$ size generated from an image of $224\times 224$ size can not process the feature map of $28\times 28$ size from an image of $448\times 448$ size. In contrast, the architecture of our S$^2$-MLP is invariant to the input scale.

\vspace{0.1in}

\subsection{Ablation study}
 Due to limited computing resources, the ablation study is conducted on ImageNet100, which is a subset of ImageNet-1K containing images of randomly selected 100 categories.
Due to the limited space, the ablation study in this section only includes that with
 the wide settings.
 We only change one hyperparameter at each time and keep the others the same as the wide settings in Table~\ref{para}.

\vspace{0.1in}
\noindent \textbf{Depth.}\; The proposed S$^2$-MLP architecture stacks $N$ S$^2$-MLP blocks. We evaluate the influence of depth  ($N$) on recognition accuracy, the number of parameters and FLOPs.   As shown in Table~\ref{depth}, as  the depth $N$ increases from $1$ to $12$, the recognition accuracy increases accordingly.  This is expected since more blocks have a more powerful representing capability. To be specific, when $N=1$, it only achieves a $56.7\%$ top-1 accuracy and $82.0$ top-5 accuracy.    In contrast, when $N=12$, it attains $87.1\%$ top-1 accuracy and $92.1\%$ top-5 accuracy. Meanwhile, the number of parameters increases from $6.5$M to $71$M and FLOPs increases from $1.3$B to $14$B  when $N$ increases from $1$ to $12$. We also observe that, when $N$ further increases from $12$ to $16$, the retrieval accuracy drops. This might be due to the over-fitting since ImageNet100 is relatively small-scale. When  training  on a huge-scale dataset such as JFT-300M,   $L=16$ might achieve higher accuracy than $L=12$. Considering both efficiency and effectiveness,  $N=12$ is a good choice. 

\vspace{0.1in}

\begin{table}[t]
\centering
\begin{tabular}{c|c|c|c|c}
\hlineB{3}
$N$  & Top-1 ($\%$)& Top-5 ($\%$) & Para.  (M) & FLOPs (B) \\ \hline
1  &   $56.7$   &  $82.0$     &  $6.5$      &   $1.3$    \\
3  &   $79.6$   &  $94.1$    &   $18$     &     $3.6$  \\
6  &   $84.6$   &   $96.0$   &   $36$     &     $7.1$  \\
12 &   $87.1$   &   $97.1$    &  $71$       &    $14$  \\
16 &   $86.3$   &   $96.9$    &   $95$     &    $19$  \\\hlineB{3}
\end{tabular}
\vspace{0.05in}
\caption{The influence of the number of blocks, $N$.}
\label{depth}\vspace{0.05in}
\end{table}

\begin{table}[t]
\centering
\begin{tabular}{c|c|c|c|c}
\hlineB{3}
$c$  & Top-1 ($\%$)& Top-5 ($\%$) & Para.  (M) & FLOPs (B) \\ \hline
$192$  &   $79.7$   &  $94.7$     &   $4.3$     &  $0.9$     \\
$384$  &   $85.3$   &  $96.6$    &    $17$    &    $3.5$   \\
$576$  &   $85.7$   &   $96.7$   &    $38$    &    $7.9$   \\
$768$ &   $87.1$   &   $97.1$    &    $71$       &    $14$  \\
$960$ &   $87.0$   &   $97.0$    &   $106$     &    $20$  \\\hlineB{3}
\end{tabular}
\vspace{0.05in}
\caption{The influence of the hidden size, $c$.}
\label{hidden}\vspace{0.05in}
\end{table}

\begin{table}[t]
\centering
\begin{tabular}{c|c|c|c|c}
\hlineB{3}
$r$  & Top-1 ($\%$)& Top-5 ($\%$) & Para.  (M) & FLOPs (B) \\ \hline
$1$  &   $86.1$   &  $96.7$     & $29$       &  $5.7$     \\
$2$  &   $86.4$   &  $96.9$    &  $43$      &  $8.4$     \\
$3$  &   $87.0$   &   $96.8$   &    $57$    &   $11$    \\
$4$ &   $87.1$   &   $97.1$    &  $71$       &    $14$     \\
$5$ &   $86.6$   &   $96.8$    &   $86$     &    $17$  \\\hlineB{3}
\end{tabular}
\vspace{0.1in}
\caption{The influence of the expansion ratio, $r$.}
\label{expan}
\end{table}

\begin{figure*}[t]
    \centering
    \includegraphics[scale=0.7]{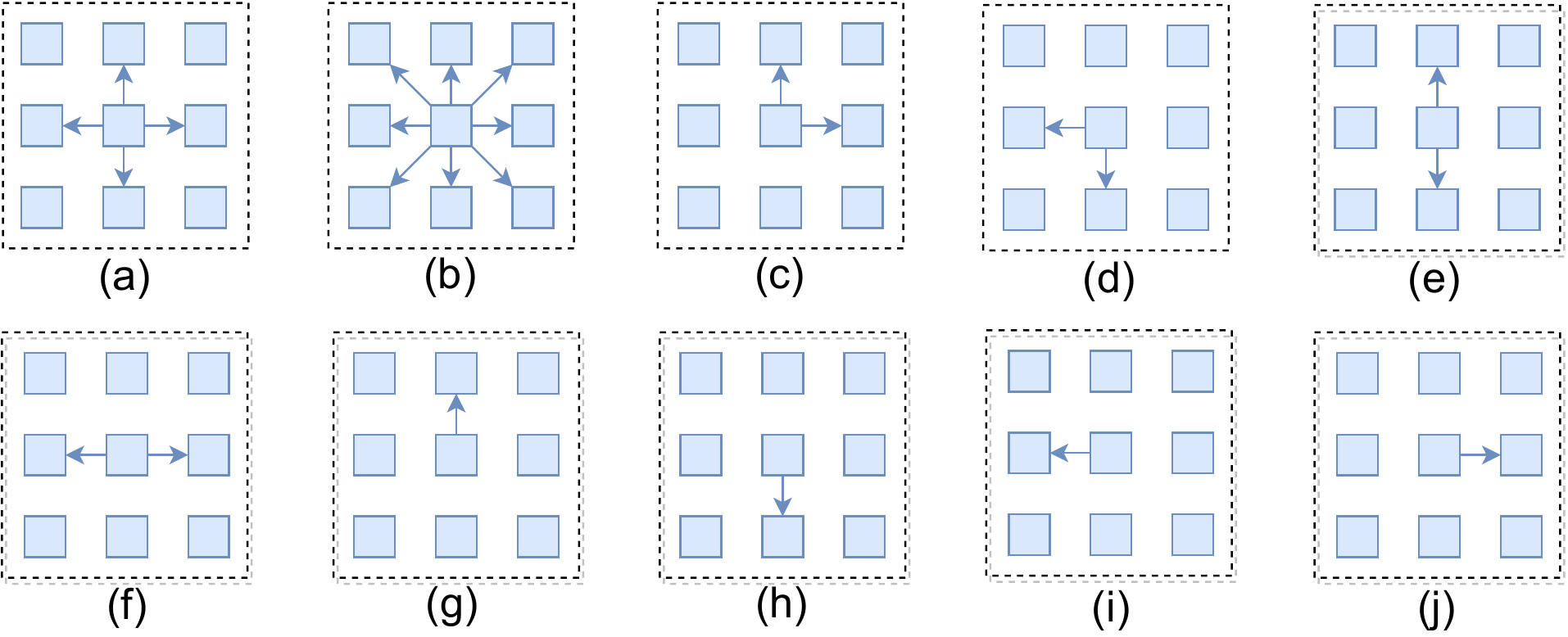}
    
    \vspace{0.1in}
    \caption{Ten different shifting settings. (a) is the default option  which shifts  channels along four directions. (b) shifts channels along eight directions. (c),(d),(e),  and (f) shift  channels in two directions. (g), (h), (i), and (j) shift channels along a single direction. }
    \label{direction}\vspace{0.1in}
\end{figure*}

\begin{table*}[t]
\centering
\begin{tabular}{c c c c c c c c c c c c}
\hlineB{3}
Settings & (a) & (b) & (c) & (d) & (e) & (f) & (g) & (h) & (i) & (j) & w/o \\ \hline
Top-1 ($\%$)   & $87.1$  & $87.0$  & $85.0$ & $85.1$  & $79.5$ & $80.5$  & $77.7$ & $77.5$ & $78.3$& $78.4$ & $56.7$\\ 
Top-5  ($\%$)  &   $97.1$   &  $97.1$  & $96.1$  & $96.2$  &  $93.1$ & $93.7$ & $92.7$ & $92.5$ & $93.4$& $93.4$ & $81.0$ \\\hlineB{3}
\end{tabular}
\vspace{0.1in}
\caption{The influence of shifting directions.}
\label{shift}
\end{table*}

\vspace{0.1in}
\noindent \textbf{Hidden size.}\; The hidden size $(c)$ in MLPs of S$^2$-MLP blocks also determine the modeling capability of the  proposed S$^2$-MLP architecture.  In Table~\ref{hidden}, we show the influence of $c$.
As shown in the table, the top-1 recognition accuracy  increases from $79.7\%$ to $87.1\%$ as the hidden size $c$ increases from $192$ and $768$, and the number of parameters increases from $4.3$M to $71$M, and FLOPs increases from $0.9$b to $14$B. The recognition accuracy saturates when $c$ surpasses $768$. Taking both accuracy and efficiency into consideration, we set $c=768$, by default.

\vspace{0.1in}
\noindent \textbf{Expansion ratio.}\; Recall that the weights of the third layer and the fourth fully-connected layer, $\mathbf{W}_3 \in \mathbb{R}^{rc\times c}$ and $\mathbf{W}_4 \in \mathbb{R}^{c\times rc}$. $r$  determines the modeling capability of these two fully-connected layers in each S$^2$-MLP block. Table~\ref{expan} shows the influence of $r$. As shown in the table, the top-1 accuracy increases from $86.1\%$ to $87.0\%$ as $r$ increases from $1$ to $3$. Accordingly, the number of parameters increases from 29M to 57M. 
But the accuracy saturates  and even turns worse when $r$ surpasses $3$. This might be due to the fact that ImagetNet100 is too small and our model suffers from over-fitting when $r$ is large.

\vspace{0.1in}
\noindent \textbf{Shifting directions.}\; By default, we split $768$ channels into four groups and shift them along four directions as Figure~\ref{direction} (a).  We also attempt other shifting settings.  (b) splits the channels into $8$ groups, and shift them along eight directions.  (c), (d), (e), and (f) split the channels into two groups, and shift them along two directions.  (g), (h), (i), and (j) shift all channels along a single direction.  In Table~\ref{shift}, we 
show the recognition accuracy of our S$^2$-MLP with shifting from (a) to (j). We also show that achieved by S$^2$-MLP without (w/o) shifting. As shown in the table, without shifting, the network performs poorly due to a lack of communications between patches. Besides, comparing (e) with (f), we discover that the horizontal shifting is more useful than the vertical shifting. Comparing (c) with (e)/(f),  we observe that shifting in two dimensions (both horizontal and vertical) will be helpful than shifting in a single dimension (horizontal or vertical). Moreover, comparing (a) and (b), we conclude that shifting along four directions is enough. Overall, the default shifting configuration, (a), the most natural way for shifting, achieves excellent performance.

\vspace{0.1in}
\noindent \textbf{Input scale.}\; The input image is resized into $W\times H$ before being fed into the network. When the patch size $p$ is fixed, the image of larger scale will generate more patches, which will inevitably  bring more computational cost. But a larger scale is beneficial for modeling  fine-grained details in the image, and generally leads to higher recognition accuracy.

\begin{table}[h]
\centering
\begin{tabular}{c|c|c|c|c}
\hlineB{3}
$W\times H$  & Top-1 ($\%$)& Top-5 ($\%$) & Para.  & FLOPs \\ \hline
$112\times 112$  &   $80.6$   &  $94.2$     &   $71$M &  $3.5$B     \\
$224 \times 224$ &   $87.1$   &   $97.1$    &   $71$M & $14$B     \\
$384 \times 384$ &   $88.2$   &   $97.6$    &   $71$M &  $31$B    \\\hlineB{3}
\end{tabular}
\vspace{0.1in}
\caption{The influence of the input image scale.}
\label{scale}
\end{table}

Table~\ref{scale} shows the influence of the input image scale. As shown in the table, when $W \times H$ increases from $112\times 112$ to $336 \times 336$, the top-1 recognition accuracy improves from $80.6\%$ to $88.2\%$, the number of parameters keeps unchanged since the network architecture does not change, and the FLOPs also increases from $3.5$B to $31$B. Note that, when the input scale increases from $224\times 224$ to $384\times 384$, the gain in recognition accuracy is not significant, but the FLOPs is doubled. Therefore, we only recommend to adopt a large-scale input if the computing resources are abundant.

\vspace{0.1in}
\noindent  \textbf{Patch size.}\; When the input image scale is fixed, the increase of patch size will reduce the number of patches. 
The larger-size patch enjoys high efficiency but is not good at capturing the fine-level details. Thus, the larger-size patches cannot achieve as high accuracy as its smaller counterparts. As shown in Table~\ref{patch}, when $p$ increases from $16$ to $32$, it reduces FLOPs from $14$B to $3.5$B. But it also leads to that the top-1 recognition accuracy drops from $87.1\%$ to $81.0\%$. Thus, we only recommend to use the larger-size patch in the case demanding fast inference.

\begin{table}[t]
\centering
\begin{tabular}{c|c|c|c|c}
\hlineB{3}
$p\times p$  & Top-1 ($\%$)& Top-5 ($\%$) & Para.  & FLOPs  \\ \hline
$32\times 32$  &    $81.0$   &   $94.6$    &   $73$M & $3.5$B   \\
$16 \times 16$ &   $87.1$   &   $97.1$    &   $71$M & $14$B    \\ \hlineB{3}
\end{tabular}
\vspace{0.1in}
\caption{The influence of the patch size.}
\label{patch}
\end{table}

\vspace{0.1in}
\section{Conclusion}
In this paper, we propose a spatial shift MLP (S$^2$-MLP) architecture. It adopts a pure MLP structure without convolution and self-attention. To achieve the communications between spatial locations, we adopt a spatial shift operation, which is simple, parameter-free, and efficient. On ImageNet-1K dataset, S$^2$-MLP achieves considerably higher recognition accuracy than the pioneering work, MLP-Mixer and ResMLP, with a comparable number of parameters and FLOPs. Compared with its ViT counterpart, our S$^2$-MLP takes a simpler architecture, with less number of parameters and FLOPs . Moreover, we  discuss the relations among the spatial shifting operation,  token-mixing MLP in MLP-Mixer, and the depthwise convolution. We discover that both token-mixing MLP and the proposed spatial-shift operation are variants of the depthwise convolution. We hope that these results and discussions could inspire further research to discover simpler and more effective vision architecture in the near future.

\vspace{0.15in}

\balance
{
\bibliographystyle{ieee_fullname}
\bibliography{egbib}
}

\end{document}